\RequirePackage[loading]{tracefnt}
\documentclass[letterpaper, 10pt, conference]{ieeeconf}   
\usepackage{standalone} 
\IEEEoverridecommandlockouts                              
\usepackage{dsfont} 
\usepackage{amsmath} 
\usepackage{amssymb}  
\usepackage{graphicx} 
\usepackage{booktabs}
\usepackage{cite}
\usepackage{macros}
\usepackage{hyperref}
\usepackage{bm}
\usepackage{tikz}
\usetikzlibrary{calc, positioning, arrows.meta}
\usepackage{pgfplots}
\pgfplotsset{compat=1.17}

\usepackage[font=normalsize]{caption}
\usepackage[font=normalsize]{subcaption}
\newcommand{\norm}[1]{\left\lVert#1\right\rVert}

\def\figWidth{0.85}

\title{\LARGE \bf Generalizability of Graph Neural Networks for\\ Decentralized Unlabeled Motion Planning}

\author{Shreyas Muthusamy$^{1}$, Damian Owerko$^{1}$, Charilaos I. Kanatsoulis$^{2}$, Saurav Agarwal$^{1}$, and Alejandro Ribeiro$^{1}$
\thanks{$^{1}$Dept. of Electrical and Systems Eng., University of Pennsylvania, USA {\tt\{muthurak, owerko,sauravag,aribeiro\}@seas.upenn.edu}}%
\thanks{$^{2}$Dept. of Computer Science, Stanford University, USA 
    {\tt charilaos@cs.stanford.edu}}
\thanks{Work was supported by the grant ARL DCIST CRA W911NF-17-2-0181.}
}%

\begin{document}

\maketitle
\pagestyle{plain}

\begin{abstract}
Unlabeled motion planning involves assigning a set of robots to target locations while ensuring collision avoidance, aiming to minimize the total distance traveled.
The problem forms an essential building block for multi-robot systems in applications such as exploration, surveillance, and transportation.
We address this problem in a decentralized setting where each robot knows only the positions of its $k$-nearest robots and $k$-nearest targets.
This scenario combines elements of combinatorial assignment and continuous-space motion planning, posing significant scalability challenges for traditional centralized approaches.
To overcome these challenges, we propose a decentralized policy learned via a Graph Neural Network (GNN).
The GNN enables robots to determine (1)~\emph{what} information to communicate to neighbors and (2)~\emph{how} to integrate received information with local observations for decision-making.
We train the GNN using imitation learning with the centralized Hungarian algorithm as the expert policy, and further fine-tune it using reinforcement learning to avoid collisions and enhance performance.
Extensive empirical evaluations demonstrate the scalability and effectiveness of our approach.
The GNN policy trained on 100 robots generalizes to scenarios with up to 500 robots, outperforming state-of-the-art solutions by 8.6\% on average and significantly surpassing greedy decentralized methods.
This work lays the foundation for solving multi-robot coordination problems in settings where scalability is important.

\end{abstract}

\section{Introduction}%
\label{sec:intro}%
A fundamental challenge in large-scale multi-robot systems is the unlabeled motion planning problem, where a group of identical robots must be assigned to a set of target locations without predetermined pairings, ensuring collision-free paths while minimizing the total distance traveled~\cite{Khan2021-dc}.
This problem is inherently complex because it combines the difficulties of the assignment problem—a combinatorial optimization challenge—with those of motion planning in continuous geometric spaces.
Efficiently solving unlabeled motion planning has significant implications for applications such as warehouse automation~\cite{Baker2007-zv}, swarm robotics~\cite{Schranz2020-uc}, search and rescue missions~\cite{Baxter2007}, and exploration~\cite{burgard-exploration}.

\begin{figure}
    \centering
    \vspace{1em}
    	\begin{tikzpicture}
        \definecolor{mBlue}{HTML}{1F77B4}
        \definecolor{mDarkRed}{HTML}{a2282f}
        \definecolor{mSteelGray}{HTML}{CACACA}
        \footnotesize
        \tikzstyle{Arc} = [thick,-{Latex[length=2mm,width=1.8mm]}]
        \tikzstyle{robot}=[circle,draw=mBlue,thick,fill=white,inner sep=0.4mm, outer sep=0, text=mBlue]
        \tikzstyle{target}=[circle,thick,draw=mDarkRed,fill=white,inner sep=0.4mm, outer sep=0, text=mDarkRed]
        \tikzstyle{arrow}=[black, -Stealth, thick]
        \begin{axis}[xmin=-0.5,xmax=5.5,ymin=-0.5,ymax=1.5, xtick distance=1, ytick distance = 1, height=2in, width=3.5in, xlabel=$x$, ylabel=$y$]
            \node (r1) at (2,1) [robot]{$1$};
            \node (r2) at (0,0) [robot]{$2$};
            \node (t1) at (3,1) [target]{$1$};
            \node (t2) at (5,0) [target]{$2$};

            \draw[arrow] (r1) -- (t1) node[pos=0.5,above]{$1$};
            \draw[arrow] (r2) -- (t2) node[pos=0.5,above]{$5$};
            \draw[arrow, dashed] (r1) -- (t2) node[pos=0.5,sloped, above]{$\sqrt{10}$};
            \draw[arrow, dashed] (r2) -- (t1) node[pos=0.5,sloped, above]{$\sqrt{10}$};
        \end{axis}
        
	\end{tikzpicture}
    \caption{Example for difference between assignment using sum of distances \(\bfP_\text{LSAP}\) (solid lines) and sum of distances squared, \(\bfP_\text{CAPT}\) (dashed lines). Agents are blue, and goals are red. \(\bfP_\text{CAPT}\) prioritizes reducing the maximum distance traveled for each agent, which increases total distance.
    }
    \label{fig:assignment_comparison}
\end{figure}
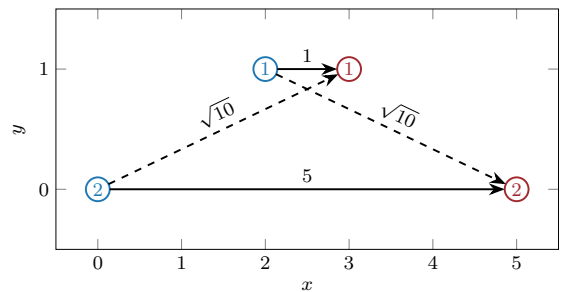

A common formulation for assigning robots to target locations is the Linear Sum Assignment Problem (LSAP)~\cite{Kuhn1955-ih}, which seeks to minimize the total distance between robots and their assigned targets.
The LSAP is well-studied, and algorithms based on the Hungarian method~\cite{Kuhn1955-ih} can solve it in polynomial time~\cite{BurkardDM12}.
However, these algorithms are centralized and have cubic complexity. Therefore, they do not scale well with large numbers of robots.
Moreover, the LSAP does not account for inter-robot collisions, which are crucial in multi-robot systems.
Approaches often rely on online replanners or potential field methods~\cite{khatib1986real}, which can prevent collisions but often disregard the global optimality of the solution.

Another approach minimizes the sum of squared distances for the assignment and then employs constant velocity trajectories so that all robots reach their respective targets simultaneously, as seen in the Concurrent Assignment and Planning of Trajectories (CAPT) algorithm~\cite{turpin-capt}.
This method ensures collision avoidance under mild restrictions on the initial configurations of the robots.
However, the solution may not be optimal in terms of the total distance traveled; for example, Figure~\ref{fig:assignment_comparison} illustrates a scenario where the CAPT solution is suboptimal.
Additionally, since CAPT also relies on the Hungarian algorithm, it inherits scalability issues with large numbers of robots.

To mitigate the issues of a centralized system, we consider the decentralized setting, where the robots have only localized information and communicate with neighboring robots.
Determining an optimal controller in a decentralized setting is difficult for many continuous-space tasks~\cite{Witsenhausen1968-wf}, leading to the development of approximate methods.
Recently, learning-based methods, in particular graph neural networks (GNNs) have been used for multi-robot systems for applications such as multi-agent pathfinding~\cite{Kipf2016-hz}, flocking~\cite{tolstaya-gnn-il}, target tracking~\cite{zhou-target-tracking}, and coverage control~\cite{gosrich-coverage-control,agarwal2024lpac}.
A common theme of these approaches is that an expert clairvoyant algorithm is used to generate a dataset for imitation learning.

Previous research has demonstrated that combining GNNs and reinforcement learning can effectively address problems that are unsolvable with imitation learning approaches alone \cite{blumenkamp-gnn-rl, Khan2021-dc}. However, reinforcement learning is known to be a sample inefficient learning algorithm~\cite{Botvinick2019-ax}. It is possible to lower the number of samples needed to train reinforcement learning, though, by referencing an optimal policy in the reward function~\cite{NEURIPS2021_e61eaa38}.
Pre-training a model with imitation learning and fine-tuning with reinforcement learning can further improve sample efficiency~\cite{ramrakhya-pirlnav}. 
To optimize the policy and obtain desired results, reward engineering has also been used to find a reward function to efficiently improve a policy~\cite{peng-deepmimic}.

In this paper, we present a learning-based approach that minimizes the total distance traveled while avoiding collisions. We focus on a decentralized setting, where robots have local information and limited communication capabilities. Under these constraints, we show that a GNN can imitate a centralized approach, outperforming decentralized baselines. Fine-tuning with reinforcement learning further improves GNN performance and provides nearly collision-free control. To our knowledge, this work is the first example of a continuous-space multi-agent system used for motion planning being trained using imitation learning and fine-tuned using reinforcement learning. Finally, we present empirical results that demonstrate the scalability of the proposed approach: a model trained on 100 robots generalizes to scenarios with up to 500 robots.

\section{Unlabeled Motion Planning}\label{sec:unlabeled_motion_planning}

Consider \(N\) agents with radius \(R\) and a set of \(N\) goal positions. We study the problem of navigating the agents from their original positions to the goal positions. There is no prior assignment of agents to goals. Hence, this problem is known as \emph{unlabeled} motion planning. To describe the problem mathematically, let the positions of the agents be
\begin{equation}\label{eq:agent_position}
    \bfX(t) = \bmat{\bfx_1(t) & \bfx_2(t) & \dots & \bfx_N(t) }^T \in \reals^{N \times 2}
\end{equation}
where \(\bfx_i(t) \in \reals^2\) is the position of the \(i^{\mathrm{th}}\) at the time \(t\). Similarly, let 
\begin{equation}\label{eq:goal_position}
    \bfG = \bmat{\bfg_1 & \bfg_2 & \dots & \bfg_N }^T \in \reals^{N \times 2}
\end{equation}
represent the goal positions.

In this article, we consider a first-order dynamical model of the agents' motion. The input to the dynamical system is each agent's velocity,
\begin{equation}\label{eq:control}
    \bfU(t) = \bmat{\bfu_1(t) & \bfu_2(t) & \dots & \bfu_N(t) }^T \in \reals^{N \times 2}.
\end{equation}
Like in Equation \eqref{eq:agent_position}, the velocity of the \(i^\text{th}\) agent is \(\bfu_i(t)\) at time \(t\). The control inputs are constrained by the maximum velocity \(u_\mathrm{max} \in \reals_+\):
\begin{equation}\label{ineq:velocity_constraint}
    \norm{\bfu_i(t)} \le u_\mathrm{max}.
\end{equation}
Therefore, we can describe the motion of the agents in discrete time as
\begin{equation}\label{eq:dynamics}
    \bfX(t+1) = \bfX(t) + \bfU(t).
\end{equation}
In Equation \eqref{eq:dynamics}, we assume the time step to be one, without loss of generality. This simplifies notation, but we discretize at a finer interval of 0.1 seconds in our implementation. 

To quantify the quality of a solution, we introduce \emph{coverage} in Equation \eqref{eq:coverage}. It is defined as the proportion of goals that have an agent within a threshold distance \(R_c\).
\begin{equation}\label{eq:coverage}
    c(t) = c(\bfX(t), \bfG) = \frac{1}{N} \sum_{n=1}^N \mathds{1}(\min_j \norm{\bfg_i-\bfx_j}_2 < R_c).
\end{equation}
We can calculate coverage by finding the closest agent to each goal and checking if the distance is less than \(R_c\). 
To be practically viable, a control policy should produce trajectories that have few or no collisions. Define \(p_i(t)\) as the number of collisions at a time \(t\) for the \(i^\text{th}\) agent,
\begin{equation}\label{eq:collisions}
    p_i(t) = \sum_{j \ne i} \mathds{1}(\norm{\bfx_i(t) - \bfx_j(t)}_2 < 2R)
\end{equation}
where \(R\) is the radius of the agents. An ideal controller would produce trajectories satisfying \(p_i(t) = 0\), but we can practically allow \(p_i(t) \approx 0\).

A potential solution to the unlabeled motion planning problem is a control \emph{policy} \(\Pi(\bfX(t), \bfG)\) such that 
\begin{equation}
    \bfU(t) = \Pi(\bfX(t), \bfG).
\end{equation}
The policy is feasible if it satisfies Inequality \eqref{ineq:velocity_constraint}. We want to find a policy that converges to the highest coverage in the shortest amount of time. We define the \emph{discounted coverage}:
\begin{equation}\label{eq:discounted_coverage}
    C(\Pi) = \bbE_{\bfX(0),\bfG}\left[ \frac{1}{\sum_{t=0}^{T} \gamma^t} \sum_{t=0}^{T} \gamma^t c(\bfX(t), \bfG) \right] ,
\end{equation}
where \(T \in \bbZ\) is the time horizon and \(\gamma \in \reals_+\) is a discount factor. To prioritize coverage in the short term, we assume \(\gamma < 1\). A lower discount factor prioritizes short-term performance. We are interested in finding a policy that maximizes the discounted coverage over a finite horizon \(T\).


A common way to solve the unlabeled motion planning problem is to first find a matching between agents and goals. 
Using the output of the linear sum assignment problem (LSAP)~\cite{Kuhn1955-ih} with the distance between agent and goals as the cost yields one potential solution. Let \(\bfD(t) \in \reals^{N \times N}\) be a matrix with entries \(D_{ij} = || \bfx_i(t) - \bfg_j ||_2\). Then, the solution to the related LSAP problem is a permutation matrix,
\begin{equation}\label{eq:assignment_lsap}
    \bfP_\text{LSAP}(t) = \argmin_{\bfP \in \mathcal{P}_N} \mathbf{1}^T(\bfP \odot \bfD(t))\mathbf{1}
\end{equation}
where \(\calP_N\) is the set of \(N \times N\) permutation matrices and \(\odot\) is the Hadamard product. Given this assignment, \(C(\Pi)\) is maximized by each agent heading towards its assigned target at maximum velocity. The resultant trajectories never intersect unless the positions and goals of two agents are collinear~\cite{turpin-capt}.

While the trajectories do not intersect, they are not guaranteed to be collision-free for agents with a non-zero radius~\cite{turpin-capt}. To address this, \cite{turpin-capt} proposes the CAPT algorithm, which guarantees collision-free trajectories. The algorithm first uses LSAP with a cost matrix \(\bfD^2(t)\), the distance \emph{squared} between the agents and goals.
\begin{equation}\label{eq:assignment_capt}
    \bfP_\text{CAPT}(t) = \argmin_{P \in \mathcal{P}_N} \mathbf{1}^T(\bfP \odot \bfD^2(t))\mathbf{1}
\end{equation}
For the dynamics in Equation \eqref{eq:dynamics}, CAPT produces a constant velocity trajectory for each agent, such that all agents reach their assigned goal at the same time. Each agent moves directly to their assigned target, but agents with a lower travel distance move at a proportionally lower speed.

We will refer to Equation \eqref{eq:assignment_lsap} as the LSAP assignment and to Equation \eqref{eq:assignment_capt} as the CAPT assignment. Note that minimizing the sum of the distances squared, as CAPT does, leads to a different assignment than minimizing the sum of the distances. Figure \ref{fig:assignment_comparison} shows a representative example where the two solutions differ. In the LSAP policy, the first agent moves a distance of \(1\), and the second agent travels a distance of \(5\). In the CAPT solution, both agents will need to travel a distance of \(\sqrt{10}\). In our numerical experiments, we show that the LSAP policy outperforms the CAPT policy in terms of discounted coverage. However, the LSAP policy does not guarantee collision-free trajectories. Therefore, we use both imitation and reinforcement learning to find a policy that is close to the LSAP solution, but avoids collisions.

\subsection{Decentralized Motion Planning}\label{sec:decentralized}

Both the CAPT and LSAP policies are centralized, meaning that they require global information about the positions of all the agents and all the obstacles. The LSAP problem also has cubic complexity in the number of robots~\cite{Edmonds1972-kp}. A decentralized solution is well-motivated to address the issues with communication and complexity. We will say that a policy is decentralized whenever the global policy can be decomposed into,
\begin{equation}\label{eq:policy_decentralized}
    \Pi(t) = \bmat{ \pi_1(t) & \pi_2(t) & \dots & \pi_N(t) }
\end{equation}
where \(\pi_i(t)\) is a local policy computed by each agent. A policy \(\pi_i(t)\) is local if each agent independently observes its surroundings and computes the policy through a series of information exchanges with nearby agents.

We assume that each agent has information about its current velocity, and can observe the relative position of the \(k\) nearest agents and the relative positions of the \(k\) closest goals. Let \({\bfo_i(t) \in \reals^{2(1+2k)}}\) be a vector concatenating these local observations. The number of observed agents and goals could vary, but this is a minor simplification.

To model decentralized communication, we consider a communication graph \(\calG = (\calV, \calE)\) with the \(i^\text{th}\) agent represented by the \(i^\text{th}\) node in the set, \(\calV = \{1,\dots,N\}\). Let \(\calN_i\) be the set of \(k\) closest agents to the \(i^\text{th}\) agent. Then, we say that whenever \(j \in \calN_i\) then there is an edge \((j,i) \in \calE\) from \(i\) to \(j\). We allow each agent to iteratively aggregate information from its neighbors \(\calN_i\).

As an example, consider the following decentralized policy, which aggregates information from a \(d\) hop neighborhood. We will call this the \(d\)-hop decentralized policy. First, each agent finds the relative positions of all agents and goals within its \(d\)-hop neighborhood in \(\calG\). This requires \(d-1\) information exchanges on \(\calG\), since each agent already observes the relative positions of its neighboring agents without communication. Then, using this \(d\)-hop information, each agent locally solves the LSAP. The agent moves toward its locally assigned agent at maximum velocity. This family of policies has several weaknesses. First, the \(d\)-hop policy does not make guarantees for the number of collisions, similar to the centralized LSAP policy. Second, the amount of information that needs to be communicated can grow exponentially as we increase the number of hops. Finally, as we increase the number of hops, the local complexity is cubic with the number of agents in the \(d\)-hop neighborhood.

\section{Approach}

We propose to use a graph neural network (GNN) to parameterize the policy. We first use imitation learning to train the GNN to approximate the centralized LSAP policy. Such a GNN is unlikely to provide a collision-free trajectory. Therefore, we fine-tune the GNN using reinforcement learning with a reward function that balances maximizing coverage and minimizing the number of collisions. 

\subsection{Architecture} \label{sec:architecture}
The GNN architecture is well-motivated. Graph convolution layers allow neighboring agents to communicate and collaborate, and as shown by~\cite{Khan2021-dc}, GNNs can leverage the locality of the motion planning problem. A GNN can also scale to larger versions of the same problem without retraining~\cite{NEURIPS2020_12bcd658}, and has the required expressive power \cite{kanatsoulis2024graph,kanatsouliscounting} to leverage the communication graph. Additionally, between each GNN layer, we incorporate local multi-layer perceptrons (MLPs) to increase the capacity of the architecture without increasing the amount of required communication.

The GNN operates over the communication graph \(\calG\), which we defined in Section \ref{sec:decentralized}. Let \(\bfS(t) \in \{0,1\}^{N \times N}\) be an adjacency matrix associated with \(\calG\). Each edge \((i,j) \in \calE\) is represented by an entry of 1 in the \(i^\text{th}\) row and \(j^\text{th}\) column of \(\bfS(t)\). Each layer in the GNN consists of a graph convolution, a pointwise nonlinearity, and a local MLP. The input to the \(l^\text{th}\) layer of the GNN is a matrix \(\bfZ^{(l-1)} \in \reals^{N \times F}\). It is first processed by a graph convolution, which aggregates information from each agent's neighborhood:
\begin{equation}\label{eq:graph_convolution}
    \hat\bfZ^{(l)} = \sigma \left( \sum_{k=0}^{K-1} \bfS^k \bfZ^{(l-1)} \bfH_k^{(l-1)} \right).
\end{equation}
In Equation \eqref{eq:graph_convolution}, \(\bfH_k^{(l-1)} \in \reals^{F \times F}\) are the parameters of the graph convolutional layer and \(\sigma(\cdot)\) is a pointwise nonlinearity. The intermediate output, \(\hat\bfZ^{(l)} \in \reals^{N \times F}\) is then processed by an MLP to obtain an output \(\bfZ^{(l)}\), which will be fed to the next layer.
\begin{equation}\label{eq:mlp}
    \bfZ^{(l)} = \bfZ^{(l-1)} + \sigma\left( \sigma\left( \hat\bfZ^{(l)} \bfW_1 \right) \dots \bfW_{L_\text{MLP}} \right)
\end{equation}
Equation \eqref{eq:mlp} describes an MLP with \(L_{MLP}\) layers, \(F\) features and \(G\) hidden features. Hence, the weight matrices are \(\bfW_1 \in \reals^{F \times G}\) in the first layer, \(\bfW_i \in \reals^{G \times G}\) in the middle layers, and \(\bfW_{L_\text{MLP}} \in \reals^{G \times F}\) in the final layer. Notice, Equation \eqref{eq:mlp} describes an MLP which locally processes information at each agent.

The inputs to the model are the local observations made by each agent, \(\bfo_i(t)\) as defined in Section \ref{sec:decentralized}. Let 
\begin{equation}
    \bfO(t) = \bmat{\bfo_1(t) & \bfo_2(t) & \dots & \bfo_N(t)}
\end{equation}
be a matrix in \(\reals^{N \times 2(1+2k)}\), representing the observations made at each agent. The output of the model is the control actions \(\bfU(t)\) as defined in Equation \eqref{eq:control}. Therefore, denote the GNN model as \(\bm\Phi\) so that
\begin{equation}
    \bfU(t) = \bm\Phi(\bfO(t), \bfS(t);\calH)
\end{equation}
where \(\calH\) is a set of all the learnable parameters. We add MLPs at the input and output of the model to allow the number of features at the input and output to differ from \(F\).


\subsection{Imitation Learning}

We use imitation learning (IL)~\cite{Hussein2018-nx} to find a decentralized GNN policy that approximates the centralized LSAP policy. This policy is described in Section \ref{sec:unlabeled_motion_planning}. We assume that we can readily sample tuples of the form \((\bfO, \bfS, \bfU^*)\) from some abstract distribution \(\calD\). Hence, imitation learning is an empirical risk minimization problem. The solution is a set of model parameters \(\calH_\text{IL}\),
\begin{equation}
    \calH_\text{IL} = \argmin_{\calH} \bbE_{\calD} \left[ ||\bm\Phi(\bfO, \bfS; \calH) - \bfU^* ||_2^2 \right],
\end{equation}\label{eq:imitation_learning}
which minimizes the expected squared error between the output of the model and the LSAP policy. The natural way to generate samples for training is to start at some initial \(\bfX(t), \bfY\) and simulate the dynamics of the simulation. As the control action, we can either use the LSAP policy, \(\bfU^*\), or the GNN policy \(\bm\Phi\). In our experiments, we found that a mix of both is effective.

The resulting model \(\bm\Phi(\bfO, \bfS; \calH_\text{IL})\) will be an approximation of the centralized policy. Since \(\bm\Phi\) is a GNN that contains local information, we may not be able to reproduce the centralized policy. Even if we could, the LSAP policy does not perform collision avoidance. Additionally, in the decentralized setting, there may be other behaviors, such as exploration, that might improve performance in terms of discounted coverage from Equation \eqref{eq:discounted_coverage}.

\subsection{Reinforcement Learning}

We use reinforcement learning (RL) to fine-tune the policy learned through imitation learning. The goal is to find a policy that attains similar coverage, but produces collision-free trajectories. Hence, we define the reward function for the \(i^\text{th}\) as,
\begin{align}\label{eq:reward}
    r_i(t) &= \exp\left(-\frac{d_i^2}{\beta^2}\right) - \alpha p_i(t)
\end{align}
where \(d_i\) is the distance between the \(i^\text{th}\) agent and its assigned goal \(\bfy_j\) based on the LSAP assignment \(\bfP_\text{LSAP}(t)\). The values \(\alpha, \beta > 0\) are scaling coefficients. This reward function encourages agents to navigate to the optimal targets while penalizing collisions.

Deep deterministic policy gradients (DDPG)~\cite{Lillicrap2016Ccw} is an off-policy RL algorithm that can handle continuous action spaces. We use Twin Delayed DDPG (TD3)~\cite{Fujimoto2018-ns}, which is a variation on DDPG that improves training stability. In DDPG and related algorithms, two models are trained simultaneously: an actor and a critic. The actor model \(\bfU = \bm\Phi(\bfO(t), \bfS(t); \calH_{A})\) is the GNN that we pretrained with imitation learning. The critic \(\bfQ(t) = \bm\Psi(\bfO(t), \bfU(t), \bfS(t); \calH_Q)\) is a similar GNN model that additionally takes the current action as an input. The models do not share parameters, and the critic will have different input and output dimensionalities in the read-in and read-out MLPs. The output of the critic, \(\bfQ \in \reals^N\) estimates the expected future reward given the observations and actions.

Since DPPG and TD3 are off-policy RL algorithms, we assume that we can readily sample tuples \((\bfO(t), \bfS(t), \bfU(t), \bfR(t), \bfO(t+1) \sim \calD\) where \(\bfR(t) \in \reals^N\) is a vector with elements \(r_i(t)\) and \({\bfO(t+1)}\) is the observation at the next time-step. We can do this by sampling trajectories. We use the output of the actor policy with additive Gaussian noise to sample the control actions.





\section{Numerical Experiments}

In this section, we evaluate the performance of the imitation learned model and analyze the impact of fine-tuning with reinforcement. We compare the performance of both models against LSAP, CAPT, and n-hop policies. To evaluate the performance, we consider both the coverage and the number of collisions. In particular, we calculate the coverage with a discount factor of $\gamma = 0.99$ and threshold distance \(R_c = 0.2\).

For training and evaluation, we simulate the dynamical system as follows. We initialize the positions of the agents and goals uniformly within a \(w \times w\) region of interest. We sample the positions so that no two agents or goals are within \(2R\) of each other with \(R = 0.05\). We assume that each agent can communicate with the \(k = 3\) nearest agents and that it can observe the same amount of nearest agents and obstacles. Given these initial conditions, we simulate the system for 20 seconds with \(T = 200\) discrete time steps. The maximum velocity of each agent is \(u_\text{max} = 0.5\) meters per second.


\subsection{Imitation Learning}

\begin{figure*}
   \centering
   \hfill
   \begin{subfigure}{0.495\linewidth}
       \centering
       \includegraphics[width=\figWidth\linewidth]{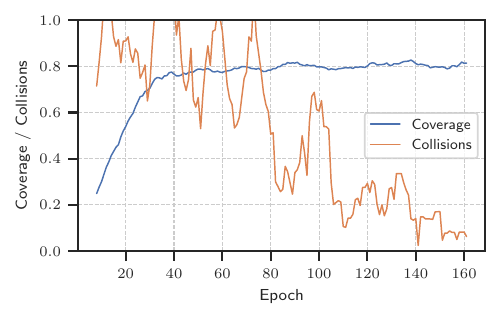}
       \caption{Imitation Learning}
       \label{fig:il_train}
   \end{subfigure}
   \hfill
   \begin{subfigure}{0.495\linewidth}
       \centering
       \includegraphics[width=\figWidth\linewidth]{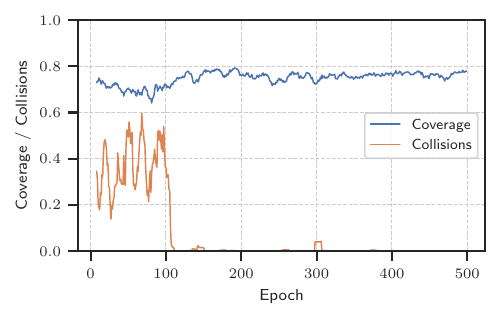}
       \caption{Reinforcement Learning}
       \label{fig:rl_train}
   \end{subfigure}
   \hfill
   \vspace{-1em}
   \caption{The average coverage \(\sum_t^T c(t) / T\) and number of collisions per agent \(\sum_t^T \sum_i^N p_i(t) / N T \) that the GNN achieves at each epoch. We run 10 simulations after each epoch and report the mean. The values are smoothed using a 9 epoch wide rolling window.}
   \label{fig:enter-label}
\end{figure*}

We use the GNN architecture described in \ref{sec:architecture}. The model is composed of \(L=5\) layers and \(K = 3\) filter taps. Each layer contains a graph convolution with feature dimension \(F = 128\) followed by an MLP with depth \(L_\text{MLP} = 3\) and \(G = 256\) hidden features. To optimize the model parameters, we use AdamW~\cite{loshchilov-adamw} a learning rate of $5 \cdot 10^{-4}$, weight decay of $10^{-8},$ and a batch size of $512$. We train the GNN using imitation learning for 161 epochs. At the beginning of each epoch, we sample 100 trajectories that are stored in an experience replay buffer with a maximum size of 100,000. During training, we use \(w = 10\) meters with \(N = 100\) agents. 

Figure \ref{fig:il_train} shows the average coverage and number of collisions per agent at different training epochs. After 40 epochs, the coverage plateaus at 0.8. The number of collisions per agent decreases over time and continues to decrease, even after the coverage plateaus. This suggests that the LSAP policy already provides a suitable solution in terms of collision avoidance, even though, as evidenced by Table \ref{tab:collisions}, it is not collision-free. 

\begin{figure}
    \centering
    \includegraphics[width=\figWidth\linewidth]{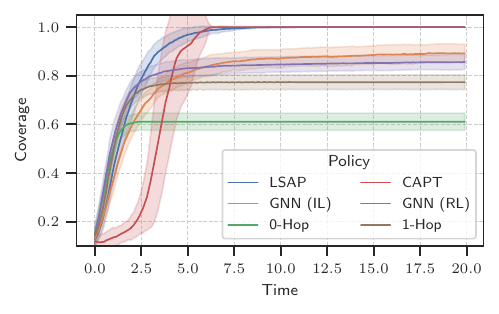}
    \caption{The coverage $c(t)$ over time for the centralized, decentralized, and GNN policies. At each time and for each policy, we plot the mean coverage from 50 simulations. The bands represent the standard deviation.}
    \label{fig:comparison}
\end{figure}

\begin{table}
    \caption{Discounted coverage \(C(\Pi)\), total collisions \(\sum_i^N \sum_t^T p_i(t)\), and total near collisions. For each policy, we report the mean values after fifty simulations. We count a near collision whenever two agents are within \(4R\).}
    \label{tab:collisions}
    \centering
    \begin{tabular}{lcrr}
        \toprule
        \textbf{Policy} & \textbf{Discounted Coverage} & \textbf{Collisions} & \textbf{Near Collisions} \\
        \midrule
        LSAP & 0.84 & 4.10 & 42.80 \\
        CAPT & 0.70 & 0.00 & 2.18 \\
        GNN (IL) & 0.72 & 45.20 & 75.90 \\
        \textbf{GNN (RL)} & \textbf{0.76} & \textbf{0.02} & \textbf{1.28} \\
        0-Hop & 0.57 & 10241.38 & 10614.20 \\
        1-Hop & 0.70 & 40.90 & 147.58 \\
        \bottomrule
    \end{tabular}
\end{table}

Figure \ref{fig:comparison} shows the average result of the policy over 50 random realizations, as well as its comparison to the centralized and decentralized policies. The figure shows that, in terms of coverage, the IL policy clearly outperformed the decentralized n-hop policies. The IL policy is dominated by the centralized LSAP policy on which it was trained. Compared to CAPT, the IL policy initially performs better but plateaus at a lower value. Table \ref{tab:collisions} summarizes the results numerically. The IL policy achieves a discounted coverage of 0.72, which is slightly better than the 0.70 that CAPT achieves. This is due to how long it takes the CAPT policy to ramp up. The IL policy performs poorly in terms of collisions, outperforming only the 0-hop policy -- under this policy, each agent travels to their closest goal.


\subsection{Reinforcement Learning}

After pre-training the GNN with IL, we use RL to fine-tune the parameters. The goals are to improve coverage and reduce collisions. Similarly to \cite{ramrakhya-pirlnav}, for the first 100 epochs, we freeze the actor model and train only the critic. The initial learning rates are \(0\) and \(10^{-4}\) for the actor and critic, respectively. Then, for the next \(50\) epochs, the learning rates are linearly interpolated to \(10^{-5}\) and \(5 \cdot 10^{-5}\), respectively. Afterward, we continue training until we reach \(500\) epochs. The effects of this schedule are clearly visible in Figure \ref{fig:rl_train}. The scaling coefficients in Equation (\ref{eq:reward}) for the policy where $\alpha = 30$ and $\beta = 0.1.$ All other hyperparameters are the same as during imitation learning.

As shown by Table \ref{tab:collisions}, the RL policy achieves an average of 0.02 collisions per simulation. This is a drastic improvement from 45.20 for the IL policy and is almost collision-free. For comparison, the LSAP policy had 4.10 collisions on average. At the same time, the discounted coverage improved by 5.6\% after fine-tuning to 0.76. This is an 8.6\% improvement over CAPT (see Figure \ref{fig:comparison}). The RL policy has the steepest initial improvement in coverage, even though its steady state performance is lower than the IL policy.

\subsection{Generalizability}

\begin{figure}
    \centering
    \includegraphics[width=\figWidth\linewidth]{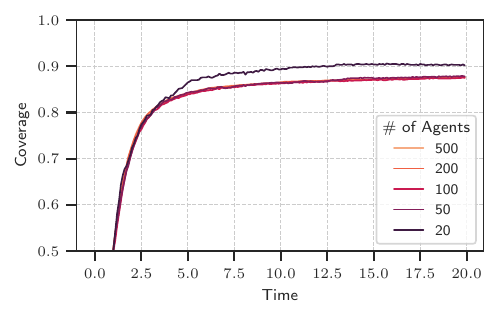}
    \caption{The coverage $c(t)$ over time for different numbers of agents, \(N\). The density of agents is constant at \(\rho = 1.0\).}
    \label{fig:generalize-scale}
\end{figure}

\begin{figure}
    \centering
    \includegraphics[width=\figWidth\linewidth]{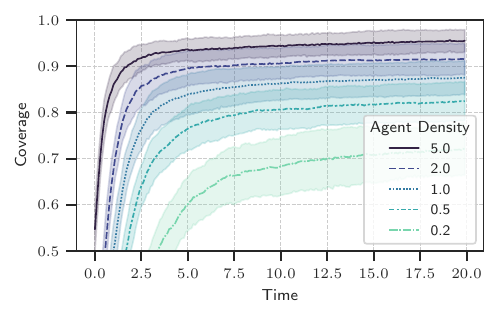}
    \caption{The coverage $c(t)$ over time for different agent densities, \(\rho\). The number of agents is constant at \(N = 100\).}
    \label{fig:generalize-density}
\end{figure}

\begin{figure}
    \centering
    \includegraphics[width=\figWidth\linewidth]{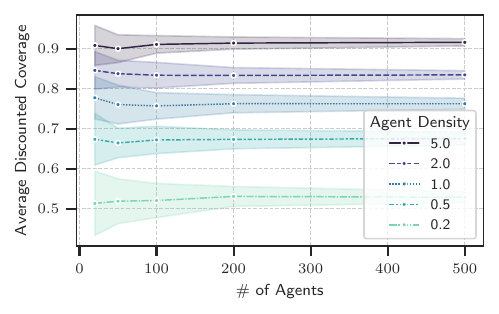}
    \caption{The discounted coverage $C(\Pi)$ for the GNN policy with $N$ agents at density \(\rho\) using the learned GNN policy in test environments with varying densities, \(\rho\). 
    }
    \label{fig:generalize-scale-density}
\end{figure}

\begin{table}
    \centering
    \caption{Effect of the number of agents \(N\) and their density \(\rho\) on the number of collisions per 100 agents.}
    \label{tab:generalize-collisions}
    \begin{tabular}{lrrrrr}
        \toprule
        \(N\)\phantom{Dummy} & 20 & 50 & 100 & 200 & 500 \\
        \(\rho\) &  &  &  &  &  \\
        \midrule
        0.2 & 152.20 & 119.56 & 99.94 & 72.45 & 57.76 \\
        0.5 & 16.40 & 10.08 & 9.24 & 8.30 & 8.28 \\
        1.0 & 0.20 & 0.28 & 0.00 & 0.20 & 0.89 \\
        2.0 & 0.40 & 0.00 & 0.50 & 0.01 & 0.04 \\
        5.0 & 0.00 & 0.04 & 0.04 & 0.04 & 0.10 \\
        \bottomrule
    \end{tabular}
\end{table}

In general, one drawback of deep learning methods is the time it takes to train.
Reusing the same model for environments with a different number of agents is desirable, particularly when we can do so without retraining \cite{Owerko23-MultiTargetTracking, Owerko23-SolvingLargeScale}.
Our proposed GNN architecture makes no assumptions about the size of the graph. Thus, we consider out-of-distribution examples with a varying number of agents and their densities. During training, \(N = 100\) agents and targets were placed in a \(w = 10\) meter wide area. This implies an average density of \(\rho = N / w^2 = 1.0\) agents per meter squared. We test the policy on environments with various number of agents \(N \in \{20, 50, 100, 200, 500\}\) and agent densities \(\rho \in \{0.2, 0.5, 1.0, 2.0, 5.0\}\). For each combination, we perform 50 simulations.

Figure \ref{fig:generalize-scale} shows the effect of changing the number of agents as their density is kept constant at \(\rho = 1.0\). The coverage is nearly identical as we scale the number of agents. The scenarios with 20 agents perform better, likely because the GNN can process the entire graph. Conversely, Figure \ref{fig:generalize-density} shows the effect of changing the density \(\rho\) while keeping the number of agents constant. Changing the density has a large impact on performance. This is partially explained by the fact that at lower densities, the average initial distance between agents and goals is higher. Specifically, since these observations contain larger values than what was seen in the training dataset, the model may not be able to generalize well to these inputs. Additionally, the lower coverage may stem from the fact that a larger distance between an agent and its corresponding goal simply means that more time is needed for the agent to reach the goal. 

The impact of changing both \(N,\rho\) is summarized by Figure \ref{fig:generalize-scale-density}, which demonstrates that the policy obtains similar discounted coverage values at the same agent density. We again note the positive correlation between density and coverage, an unintended but not necessarily harmful consequence of the GNN architecture and the nature of the problem. The impact on the number of collisions is summarized by Table \ref{tab:generalize-collisions} -- we normalize the number of collisions by \(100 / N\). As we deviate from the trained scenario, the collision frequency increases, especially for densities lower than one. However, in many scenarios, the frequency of collisions remains close to zero. For comparison, recall that we previously observed \(4.10\) collisions per 100 agents for the LSAP policy. Overall, the GNN policy can generalize most scenarios in terms of both coverage and collision avoidance.

\section{Conclusions}
The paper addressed the multi-agent unlabeled motion planning problem with the following constraints. First, the agents only perceived nearby agents and goals. Second, the agents had limited communication capability.
We proposed to use a GNN to parameterize the control policy.
Model training consisted of two phases: \emph{(i)} imitation learning produced a policy that approximated a centralized policy, and \emph{(ii)} reinforcement learning fine-tuned the policy to improve coverage and reduce collision frequency.
The final policy outperformed baseline decentralized algorithms.
Extensive empirical results established that the GNN policy is scalable to larger teams of agents.
The policy noticeably improves performance as agent density increases despite being trained with lower agent density.
Future work could involve studying the problem in an obstacle-rich environment and performing real-world experiments.


\clearpage
\bibliography{IEEEabrv,bib/settings,bib/sources}

\end{document}